\documentclass[conference]{IEEEtran}
\IEEEoverridecommandlockouts
\usepackage{cite}
\usepackage{amsmath,amssymb,amsfonts}
\usepackage{algorithmic}
\usepackage{graphicx}
\usepackage{textcomp}
\usepackage{xcolor}
\usepackage{url}
\def\BibTeX{{\rm B\kern-.05em{\sc i\kern-.025em b}\kern-.08em
    T\kern-.1667em\lower.7ex\hbox{E}\kern-.125emX}}

\usepackage{import} % to use \import{../../../content-tex/}{abstract}
\usepackage{todonotes} % USAGE: \todo[inline]{is there a better example from X topic?}
\usepackage{lipsum}  % USAGE \lipsum[2-4]
\newcommand{\etal}{\textit{et al. }} % \etal

\begin{document}

\title{
% Diversity, Equity, and Inclusion of Artificial Intelligence and Robotics for children %Fri  7 Jan 23:59:42 GMT 2022
%Piloting Inclusive Workshops of Artificial Intelligence and Robotics for children %Sat  8 Jan 00:12:35 GMT 2022
%Piloting Inclusive Artificial Intelligence and Robotics for Children %Sat 15 Jan 08:44:06 GMT 2022
%Piloting Diversity and Inclusion in Artificial Intelligence and Robotics for Children %Sat 15 Jan 10:33:54 GMT 2022
Piloting Diversity and Inclusion Workshops in Artificial Intelligence and Robotics for Children %Sun 16 Jan 13:14:45 GMT 2022
}

\author{

\IEEEauthorblockN{
    %1\textsuperscript{st} 
    A. Badillo-Perez, D. Badillo-Perez, D. Coyotzi-Molina, D. Cruz, R. Montenegro, L. Vazquez and M. Xochicale
    }
\IEEEauthorblockA{
    %\textit{dept. name of organization (of Aff.)} \\
    \textit{air4children: Artificial Intelligence and Robotics for Children}\\
    Xicohtzinco, M\'exico \\
    air4children@gmail.com
}

}

% \author{\IEEEauthorblockN{1\textsuperscript{st} Diego Coyotzi-Molina}
% \IEEEauthorblockA{\textit{dept. name of organization (of Aff.)} \\
% \textit{name of organization (of Aff.)}\\
% City, Country \\
% email address or ORCID}
% \and
% \IEEEauthorblockN{2\textsuperscript{nd} Rocio Montenegro}
% \IEEEauthorblockA{\textit{dept. name of organization (of Aff.)} \\
% \textit{name of organization (of Aff.)}\\
% City, Country \\
% email address or ORCID}
% \and
% \IEEEauthorblockN{3\textsuperscript{rd} Donato Perez-Badillo}
% \IEEEauthorblockA{\textit{dept. name of organization (of Aff.)} \\
% \textit{name of organization (of Aff.)}\\
% City, Country \\
% email address or ORCID}
% \and
% \IEEEauthorblockN{4\textsuperscript{th} Dago Cruz}
% \IEEEauthorblockA{\textit{dept. name of organization (of Aff.)} \\
% \textit{name of organization (of Aff.)}\\
% City, Country \\
% email address or ORCID}
% \and
% \IEEEauthorblockN{5\textsuperscript{th} Antonio Perez-Badillo}
% \IEEEauthorblockA{\textit{dept. name of organization (of Aff.)} \\
% \textit{name of organization (of Aff.)}\\
% City, Country \\
% email address or ORCID}
% \and
% \IEEEauthorblockN{6\textsuperscript{th} Leticia Vazquez}
% \IEEEauthorblockA{\textit{dept. name of organization (of Aff.)} \\
% \textit{name of organization (of Aff.)}\\
% City, Country \\
% email address or ORCID}
% \and
% \IEEEauthorblockN{6\textsuperscript{th} Miguel Xochicale}
% \IEEEauthorblockA{\textit{dept. name of organization (of Aff.)} \\
% \textit{name of organization (of Aff.)}\\
% City, Country \\
% email address or ORCID}
% }

\maketitle

\begin{abstract}
In this paper, we present preliminary work from a pilot workshop that aimed to promote diversity and inclusion for fundamentals of Artificial Intelligence and Robotics for Children (air4children) in the context of developing countries.
Considering the scarcity of funding and the little to none availability of specialised professionals to teach AI and robotics in developing countries, we present resources based on free open-source hardware and software, open educational resources, and alternative education programs.
% in order to promote diversity and inclusion on teaching AI and Robotics for children.
That said, the contribution of this work is the pilot workshop of four lessons that promote diversity and inclusion on teaching AI and Robotics for children to a small gender-balanced sample of 14 children of an average age of 7.64 years old.
%of the workshop including: (a) motivation and applications, (b) teaching fundamentals, (c) linking sensor and human body anatomy with robots and (d) applications and showcase presentations.
We conclude that participant, instructors, coordinators and parents engaged well in the pilot workshop noting the various challenges of having the right resources for the workshops in developing countries and posing future work.
The resources to reproduce this work are available at \url{https://github.com/air4children/hri2022}.
\end{abstract}


%%%%%%%%%%%%%%%%%%%%%%%%%%%%%%%%%%%%%%%%%
%%%%%%%%%%%%%%%%%%%%%%%%%%%%%%%%%%%%%%%%%
\section{Introduction}
%Guarantee security, accessibly and human dignity are considered the pillars for inclusivity.
Accessible and affordable technology in conjunction with open educational resources can promote equal opportunities for childhood education \cite{yoshie2021-unesco}.
However, teaching state-of-the-art technologies such as Artificial Intelligence and Robotics, AIR, is a current challenge for low-income and often politically or culturally marginalized countries.
Additionally, creating the right environment to promote inclusivity and diversity to teach AIR has been little investigated. 
Astobiza \etal 2019, for instance, reported the need of collaborations between industry and a multidisciplinary group of researchers to address concerns on the paradigm of inclusivity in robotics \cite{MonasterioAstobiza2019}.
In that sense, Astobiza \etal suggested that inclusive robotics should be based on two points: "1) they should be easy to use, and 2) they must contribute to making accessibility easier in distinct environments" \cite{MonasterioAstobiza2019}.
Peixoto \etal in 2018 reported the use of robots as tool to promote diversity leading to improve competences in communication, teamwork, leadership, problem solving, resilience and entrepreneurship \cite{PeixotoCastro2018, PeixotoGonzalez2018}. 
Recently, Pannier \etal in 2020 pointed out the challenges of increasing the  participation of women and underrepresented minorities in the areas of Mechatronics and Robotics Engineering as well as the creation of community of educators to promote diversity and inclusion \cite{Pannier2020}.
Similarly, Pannier \etal mentioned that the prevalence of free and open-source software and hardware made mechatronics and robotics more accessible to a diverse group of population.
Pannier \etal also touched on the importance of offering workshops to different range of underrepresented students leading to inspire other programs and to create outreach activities for students, trainers and workshops \cite{Pannier2020}.
In March 2021, we introduced air4children, Artificial Intelligence and Robotics for Children, as a way (a) to address aspects for inclusion accessibility, equity and fairness and (b) to create affordable child-centred materials in AI and Robotics (AIR) in developing countries \cite{montenegro2021air4children}. 
That said, in this work, we are addressing the challenges of piloting and organising workshops in the context of communities in developing countries where little to none is known about the demographics, education levels and socio-economical factors that impact on teaching AIR.  
For instance, considering the town of Xicohtzinco, Tlaxcala M\'exico as our case study, where Xicohtzinco has a total population of 14,197 (6762 males and 7435 females) \cite{inegi2022} and 19 schools including seven kindergartens (3 public and 4 private), seven primaries (4 public and 3 private), four secondaries (2 public and 2 private) and one public hight-school \cite{siged2022}.
However, neither the census \cite{inegi2022} nor the education information site \cite{siged2022} provide further information about teaching technological subjects in AI and Robotics.
% NOT QUITE SURE ABOUT THIS SPECULATIONS:
%What it is not known from the 2020 census is the education levels of the population but our anecdotal experience and from current habitants can tell that the population is a mixture of young professionals and working class population.
That said, we hypothesised that piloting workshops of air4chidren in a town such as Xicohtzinco might led us to have better understanding of the needs and challenges of promoting diversity and inclusion considering state-of-the-art technologies with open education resources.

This short paper is organised as follows:
Section II presents resources to promote diversity and inclusion in AI and Robotics for children.
Section III presents the design of workshops for children from 6 to 8 years old.
Section IV presents outcomes of a four lessons pilot workshop for 14 children including the engagement of instructors and coordinators. 
We present results of the workshops and finalise it with conclusions, limitations and future work.

%BLURS
% in Artificial Intelligence and Robotics (AIR).
% to encourage children to discover and increase their interest in 
% to promote of air4children and to create the inclusive and diversity environments might led us to have better understanding of the current needs and might provide with further evidence 
% of the real needs of the community. 
%we design and pilot workshops of air4children to test how children of different ages and genders and instructors engage to create an environment of diversity and inclusion. 

%%%%%%%%%%%%%%%%%%%%%%%%%%%%%%%%%%%%%%%%%
%%%%%%%%%%%%%%%%%%%%%%%%%%%%%%%%%%%%%%%%%
\section{Resources to promote Diversity and Inclusion in AI and Robotics for children}

\subsection{Free and open-source software, open-source hardware and open educational resources}
In March 2021, we presented examples to create educational resources aimed to be "affordable, educational and fun", such examples are (a) Otto DIY -- an educational open source robot and (b) JetBot platform -- open source educational robot to create new AI projects \cite{montenegro2021air4children}.
Similarly, considering Open Educational Resources (OER) which aim to provide "teaching and learning materials that are available without access fees" seems to be a right direction to afford innovation through OER-enabled pedagogy \cite{Clinton-Lisell2021}.
However, Wiley \etal in 2014 contrasted positives and negatives of OERS where for instance of the benefits of OERs is to make course development process quicker and easier but also highlighting the challenges of making OER material for people easier to find but with the challenge of making financially self-sustainable programs among many other difficulties \cite{Wiley2014}.
Hence, in this work, we consider Otto Humanoid as a good option because of its affordability with a cost of 200 EUROS, the block diagram programming interface, the multiple sensors and actuators (servos and LCD matrix display) aligned with open-source software and hardware and OER principles \cite{OttoDIY:2016}.

\subsection{Ensuring education and Inclusive Learning}
Recently, Opertti \etal in 2021 discussed ideas in the forum "Ensuring education and Inclusive Learning for Educational Recovery 2021" \cite{opertti2021-unesco}.
Such ideas to ensure and inclusive learning are summarised as:
(1) personalisation of education including the recognition of specific learning expectations and needs,
(2) designing inclusive, emphatic and participatory curriculum for an plural and open participation of a diversity of actors and institutions,  
(3) appropriation of technology as a community resource to strength ties between students, educators, families and communities,  
(4) empowering knowledge, learning, collaboration, trust and listening among peers, and
(5) the visualisation of schools as lifelong learning spaces.
Therefore, such ideas would help to promote diversity and inclusion of teaching AI and Robotics to children.

\subsection{Alternative education programs with new technologies}
Alternative education programs such as Montessori, Waldorf and Regio Emilia considers children as active authors of their own development \cite{edwards2002, MontessoriBOOK1969}.
%These programs have been well adopted internationally; however Edwards \etal pointed out the schools deriving from the same philosophy might also need to observe teacher-child interactions, its environments and interview to the past and present parents and children \cite{edwards2002}.
In the last 5 years such programs are starting to include topics on AI, robotics and computational thinking into their curriculum \cite{elkin2014, Aljabreen2020}.
For instance, Aljabreen pointed out the adoptions of new technologies and how early child education is re-conceptualised \cite{Aljabreen2020}. 
Elkin et al. in 2014 explored the how robots can be used in the Montessori curriculum \cite{elkin2014}.
Similarly Elkin et al. posed the question on the revision of new curriculums that include technology should not deviate from the main purpose of the Montessori classroom \cite{elkin2014}.
Drigas and Gkeka in 2016 reviewed the application of information and communication technologies in the Montessori Method, mentioning the Manipulatives, as objects to develop motor skills or understand mathematical abstractions, are based on cultural areas, language, mathematics and sensoria but little to none on technological areas \cite{DrigasGkeka2016}.
Drigas and Gkeka reviewed Montessori materials of the 21st century where interactive systems with sounds and lights, touch application to enhance visual literacy or the development of computational thinking and constructions of the physical world \cite{DrigasGkeka2016}.
These indicate that the incorporation of such manipulatives with the use of robotics might led to reach scenarios to explore motor skill development, visualisation and computational thinking. 
Recently, Scippo and Ardolino reported a longitudinal study of the use of computational thinking in five years participants of primary school in a Montessori school \cite{ScippoArdolino2021}
Scippo and Ardolino pointed out the importance of alignment of the Montessori material with the computational thinking activities. 
That said, previous authors stated various challenges on the incorporation of new technologies into their curriculum posing more questions on creating curriculums that should be more accessible to a diverse group of population as it has been done in other areas such as the case of open educational resources.

%%%%%%%%%%%%%%%%%%%%%%%%%%%%%%%%%%%%%%%%%
%%%%%%%%%%%%%%%%%%%%%%%%%%%%%%%%%%%%%%%%%
\section{Designing Diversity and Inclusion Workshops}
To design diversity and inclusive workshops we considered: (a) free and open-source software, open-source hardware and open educational resources, (b) Ensuring education and Inclusive Learning, and (c) alternative education programs with new technologies.
That said, with the combination of such resources, we proposed a four lesson workshop with three-fold aims:
(a) to promote  diversity of and inclusion to children to teach AI and Robotics with recreational and engaging activities,
(a) to encourage children to discover and increase their interest in AI and Robotics with open source hardware and software, and  
(c) to develop  the Montessori concept of 'concrete to abstract' to make abstract concepts clearer with hands-on learning materials.

\begin{figure}[t]
    % \centerline{\includegraphics[width=\linewidth]{figures/curriculum.png}}  %%OVERLEAF
    %\centerline{\includegraphics[width=\linewidth]{curriculum-design/versions/drawing-v04.png}}  %%GITHUB
    \centerline{\includegraphics[width=\linewidth]{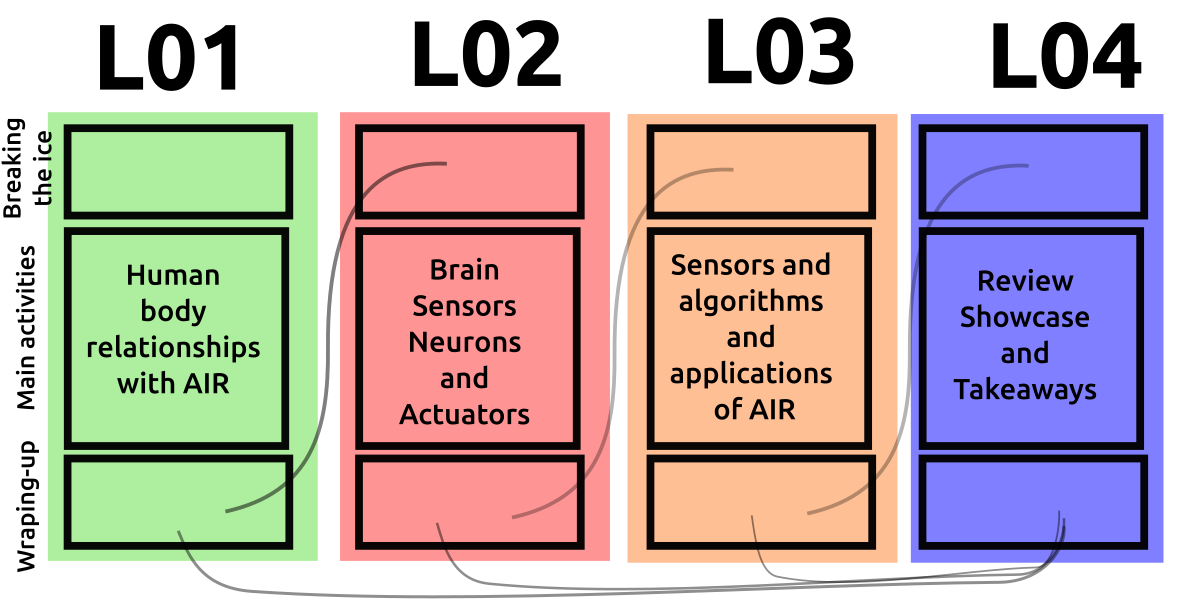}}  %%ARXIV
    \caption{Curriculum of the pilot workshop with four lessons. 
    Lesson 01 introduce the course (L01), 
    lesson 02 provides the basics of anatomy (L02), 
    lesson 03 covers algorithms (L03), and 
    lesson 04 wrap-up and showcase the project of children (L04).
    The arrows in the figure illustrate the connection from the final to initial part of each lesson and how all the lessons where connected to the final section of the workshop.
    }
    \label{fig:curriculum}
\end{figure}

\paragraph{Lesson 01: Breaking the ice and motivations} 
The educational goal for this lesson was to develop the children’s curiosity about AI and Robotics while emphasizing the importance of interpersonal connections that will evolve into a collaboration work for the following lessons.
That said, this lesson started with a recreational activity where each student and teacher introduced themselves with name, favorite food and a superpower or ability that we would like to have and that was related to robots. 
This lesson also covers basic concepts and examples of AI and Robotics in different fields and daily life, how the brain works and how the human senses and body parts relates to the way a robot is built and how it works and perform activities.

\paragraph{Lesson 02: Human senses and coding my first robot} 
The main purpose of this lesson was to understand fundamentals of Robotics. 
Children began to work with more abstract concepts, developing problem-solving skills as well as cooperatively working relationships. 
The first activity, outside of the classroom, was a true or false game, were the teacher told a sentence about AI and Robotics, and the children jumped in front of a rope if the sentence was true, or if the sentence was false, the kids jumped behind of the rope.
In the second activity, the instructor explain about the human senses and their relationship with inputs and outputs. After that, instructors explain examples of sequences and codes. 
In the last activity, participants were asked to sort out tangram in a group in which a leader of the group provide instructions to the team-mates as an analogy of coding a robot with algorithms.

\paragraph{Lesson 03: Playing with reaction-action activities} 
The educational goal of this lesson is to cover the concept of the effect of causes and consequences with daily life examples and the computational thinking of robots.
Hence, this lesson start with a match game consisting of figures and shadow’s figures where participants develop their comparing skills to match similar or different robots.
This lesson covered points on how robot works with sensors, processors, actuators and programming. 
The “find the effect” activity was also introduced where participants have to relate pictures of cause and consequences, for example “the cause is the rain and the consequently is a rainbow”.
Afterwards, we worked with the Otto humanoids in which we programmed the sensor presence for that robot moves, dances and that emit texts with the 8x8 matrix.

\paragraph{Lesson 04: Develop your own AIR}
The four lesson aimed to summaries what was covered in the previous lessons emphasising the relationship of the human body anatomy (brain, neurons and body parts) with humanoid robots (computer, sensors and actuators).
This lesson covered real-word application of AI and Robotics including medicine, spacial robotics and smart cities. 
Three projects were prepared to be introduced to each team in which every participant have a role. 
Each team prepare a short speech of their application using AI and Robotics. 

See figure \ref{fig:curriculum} that illustrates four lessons of the workshops.

%%%%%%%%%%%%%%%%%%%%%%%%%%%%%%%%%%%%%%%%%
%%%%%%%%%%%%%%%%%%%%%%%%%%%%%%%%%%%%%%%%%
\section{Piloting Diversity and Inclusion Workshops}
\begin{figure}[tbp]
    % \centerline{\includegraphics[width=\linewidth]{figures/workshop.png}} %%OVERLEAF
    %\centerline{\includegraphics[width=\linewidth]{piloting-workshops/versions/drawing-v00.png}} %%GITHUB
    \centerline{\includegraphics[width=\linewidth]{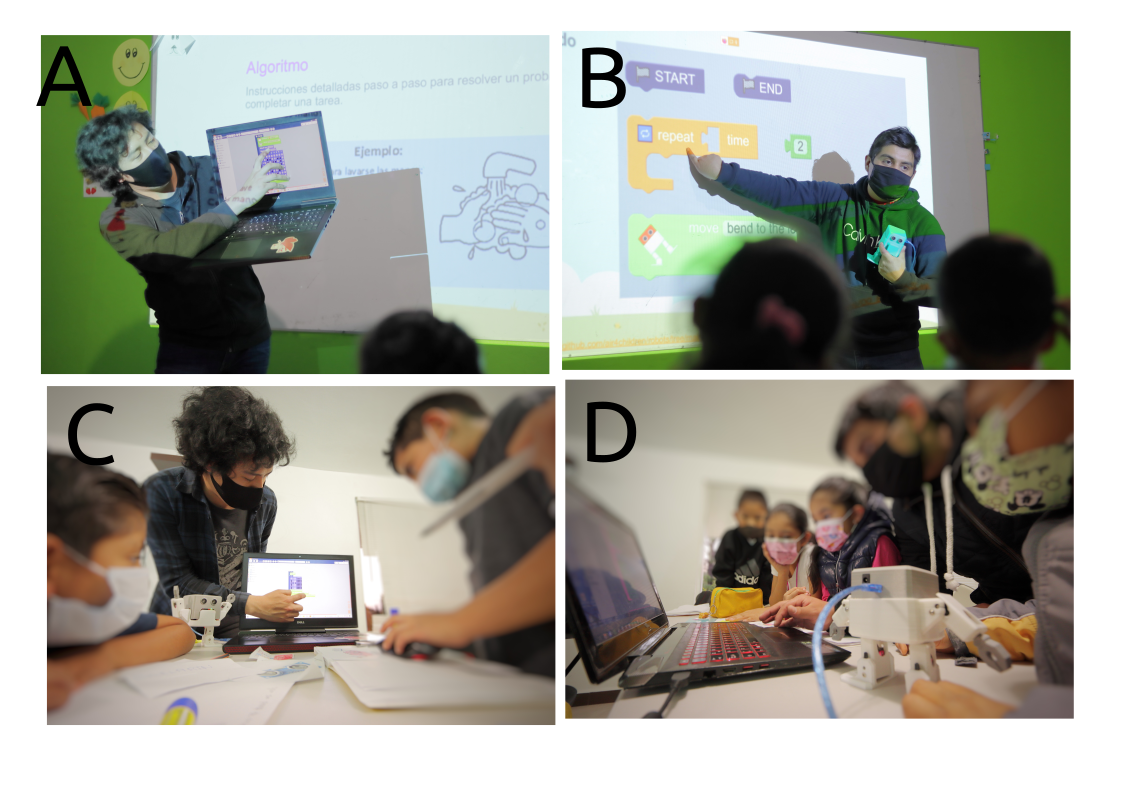}} %%ARXIV
    \caption{
        Instructors demonstrating fundamentals of AI and Robotics (A, and B). 
        Children engaging with classmates, robots and instructors (B, and C).
        }
    \label{fig:pilot}
\end{figure}

To pilot the four lesson workshop, we invited 14 participants (6 female and 8 male) with range of age from 6 to 11 years old (average age of 7.64) (Figure \ref{fig:pilot}). 
For the workshop, three instructors with three years of experience in teaching and two coordinators with ten years of teaching experience volunteered to deliver four lessons of 90 minutes in the workshop (as shown in the proposed curriculum Fig \ref{fig:curriculum}).
During the initial three lessons of the workshop, children incorporated the gained knowledge to relate fundamental human body anatomy (brain, neurons, body and senses) to robot parts (microcontroller, motors, sensors) based on open source hardware and software.
In the final lesson, children showcased their final work promoting a sense of achievement in the children working not only with their mind but also with their social emotional well-being. 
In all lessons, instructors encouraged and engaged every participant for individual and group activities.

We however noted that each lessons was originally planned to be 90 minutes of length and we did not consider breaks nor the participant's energy levels to which in the second to four lesson a 15 minutes break was incorporated.
Additionally, we piloted surveys to (a) children with ten questions about their understanding and feelings towards different type of robots and (b) to parents with 30 questions about their understanding of AI and robotics and how parents were aware of  technological advances in AI and Robotics.
Although the aim of the surveys was not to be reported but only to understand how participants and parents feel about being surveyed and how the logistics of surveys would be followed with more participants.
That said, we noticed that participants require support as few participants were not familiar with reading surveys to which the content of 10 questions was spread into five questions into two sessions.
On other hand, parents felt that surveys were lengthy, taking more than 60 minutes, and we also realised that a paper-based survey require more work as scans and transcripts are required.

% \begin{table}[htbp]
%     \caption{Table Type Styles}
%     \begin{center}
%     \begin{tabular}{|c|c|c|c|}
%     \hline
%     \textbf{Table}&\multicolumn{3}{|c|}{\textbf{Table Column Head}} \\
%     \cline{2-4} 
%     \textbf{Head} & \textbf{\textit{Table column subhead}}& \textbf{\textit{Subhead}}& \textbf{\textit{Subhead}} \\
%     \hline
%     copy& More table copy$^{\mathrm{a}}$& &  \\
%     \hline
%     \multicolumn{4}{l}{$^{\mathrm{a}}$Sample of a Table footnote.}
%     \end{tabular}
%     \label{tab1}
%     \end{center}
% \end{table}

%%%%%%%%%%%%%%%%%%%%%%%%%%%%%%%%%%%%%%%%%
%%%%%%%%%%%%%%%%%%%%%%%%%%%%%%%%%%%%%%%%%
\section{Conclusions, limitations and future work}
In this paper, we posed the challenges of promoting Diversity and Inclusion to teach "Artificial Intelligence and Robotics for Children" in developing countries, considering resources of open-source hardware and software and principles of Montessori education. 
We think that the goals of each lessons were intended to remain beyond the learning of a single concept but to contribute to develop skills of inclusion and diverstiy that children can take and apply to other areas in their life.
That said, for the pilot of the workshop, we considered a small sample of 14 children of an average age of 7.64 years old from the town Xicohtzinco Tlaxcala M\'exico. 
The workshops were free of cost as a way encourage participation and inclusion of anyone. 
During the pilot workshop, children were enthusiastic about learning the fundamentals for AIR by coding, designing and playing with open-sourced robots. 
The instructors embraced the different set of skills each child had by working in small groups and supporting the students during all the activities. 
However, we noted that grouping children of four participants with one instructor was not creating an engaging experience as each group has only one robot and one computer and the space and number of participants was leaving sometimes one participant outside of the reachable robot-computer setup.

In terms of limitations, the pilot surveys only helped us to identify the gender and age of the participants and no other insights such as needs of the target group were considered.
Similarly, this work did no consider metrics to quantify the impact of the workshop but to identify the needs of the workshop that might be addressed in future work.
The workshops were free of cost but no sustainable model was considered for this pilot experiment.

As a future work, we are planing to run another pilot in late 2022 or early 2023 with more lessons and perhaps more participants considering the addition of a study design and to run a pre-survey to identify the needs of participants of the workshops.
For the curriculum of the workshops, we are planning to improve the activities to be more engaging, diverse and inclusive and to provide further evidence on how alternative education programs (e.g. Montessori, Waldorf, Regio Emilia \cite{edwards2002}; and "synthesis program"  \cite{synthesis2022}) with new technologies might lead to potential new avenues of inclusivity and diversity.
%We will be discussing the incorporation of the "synthesis program" which aim is to cultivate the student voice, strategic thinking and collaborative problem solving skills \cite{synthesis2022}.
%started in 2014 with Josh Dahn and support of Elon Musk, 

\section*{Acknowledgment}
To Marta P\'erez and Donato Badillo for their support in organising the pilot of the workshops.
To Rocio Montenegro for her contributions with the design of the Montessori curriculum for the workshops.
To Donato Badillo Per\'ez, Antonio Badillo Per\'ez and Diego Coyotzi Molina for volunteering as instructors of the workshops.
To Leticia V\'azquez for her support with the logistics and feedback to improve the workshops.
To Adriana P\'erez Fortis for her contributions and discussion to prepare draft pilot surveys for the parents and children. 
To El\'ias M\'endez Zapata for his support and feedback on the hardware design of the robot.
To Dago Cruz for his feedback and discussions on the design of the workshops.
To Angel Mandujano, Elva Corona and others who have contributed with feedback and support to keep AIR4children project alive

\section*{Contributions}
Antonio Badillo-Per\'ez: Contributing to design and write up of lesson 02.
Donato Badillo-Per\'ezo: Contributing to design and write up of lesson 01.
Diego Coyotzi-Molina: Contributing to design and write up of lesson 03.
Dago Cruz: Contributing to proofreading, edition and feedback.
Rocio Montenegro: Contributing to the write up of designing and piloting workshops. 
Leticia V\'azquez: Write up and refinement of the conclusions.
Miguel Xochicale: Contributing to create the open source and reproducible workflow, drafting, write-up, edition, and submission of the paper. 

\bibliographystyle{IEEEtran}
% \bibliography{references} %%OVERLEAF
%\bibliography{../references/references} %%GITHUB
%\bibliography{../../references/references} %%ARXIV
% Generated by IEEEtran.bst, version: 1.14 (2015/08/26)

 %% uncomment for arxiv version

\end{document}